\documentclass[letterpaper, 10 pt, conference]{ieeeconf}  

\IEEEoverridecommandlockouts                              

\usepackage{amssymb,amsmath}
\usepackage{booktabs}
\usepackage{graphicx}
\usepackage{graphics}
\newcommand{\etal}{\textit{et al.}}

\usepackage{multicol,multirow}
\usepackage{verbatimbox}

\usepackage{hyperref}
\hypersetup{bookmarksopen,bookmarksnumbered,
pdfpagemode=UseOutlines,
colorlinks=true,
linkcolor=blue,
anchorcolor=blue,
citecolor=blue,
filecolor=blue,
menucolor=blue,
urlcolor=blue
}

\title{\LARGE \bf
LiDAR-based Crowd Navigation with
Visible Edge Group Representation
}

\author{Allan Wang$^{1}$ and Aaron Steinfeld$^{2}$ 
\thanks{*This work was supported by grants from the National Science Foundation (IIS-1734361), National Institute on Disability, Independent Living, and Rehabilitation Research (NIDILRR 90DPGE0003), and Office of Naval Research (ONR N00014-18-1-2503).}
\thanks{$^{1}$Miraikan, the National Museum of Emerging Science and Innovation, Tokyo, Japan. {\tt\small allan.wang@jst.go.jp} %
}
\thanks{$^{2}$the Robotics Institute, Carnegie Mellon University, Pittsburgh, USA. 
}
}

\begin{document}

\maketitle
\thispagestyle{empty}
\pagestyle{empty}

\begin{abstract}

Robot navigation in crowded pedestrian environments is a well-known challenge and we explore the practical deployment of group-based representations in this setting. Pedestrian groups have been empirically shown to enable a mobile robot's navigation behavior to be safer and more social. However, existing approaches either explored groups only in limited scenarios with no high-density crowds or depended on external detection modules to track individuals, which are prone to noise and errors due to occlusions in crowds. We show that group prediction accuracy affects navigation performance only marginally in crowded environments. Based on this observation, we propose the visible edge-based group representation. We additionally demonstrate via simulation experiments that our navigation framework, integrated with the simplified group representation, performs comparatively in terms of safety and socialness in dense crowds, while achieving faster computation speed. Finally, we deploy our navigation framework on a real robot to explore the benefits of practically deploying group-based representations in the real world.

\end{abstract}

\section{Introduction}

With the increasing popularity of robots in public spaces, research on social navigation, or how a robot should navigate in pedestrian-rich environments, has recently gained traction and faced new challenges~\cite{ singamaneni2024survey}. On the planning side, researchers have agreed that treating humans as dynamic obstacles is not sufficient in highly crowded environments~\cite{trautman2010unfreezing}, and how to model inter-agent interactions is identified as a core research challenge~\cite{mavrogiannis2023core}. On the perception side, robots also face difficulties identifying and tracking humans in dense environments because of heavy occlusions~\cite{wang2024tbd}, and popular human tracking models such as Bytetrack~\cite{zhang2022bytetrack} still cannot track occluded pedestrians reliably.

Recent works have deployed certain strategies to address the challenges. One common strategy is to integrate a control framework with a prediction model~\cite{heuer2023proactive, le2024social, mavrogiannis2017socially, trautman2015robot}. The prediction model predicts the interactions among individuals, and the control module plans collision-free paths. Another mainstream strategy is to leverage Deep Reinforcement Learning (DRL)~\cite{chen2017socially, liu2023intention, wang2023navistar, wu2023risk}. In these methods, the pedestrian interactions are implicitly captured by latent representations from an encoder or directly by the policy. While these strategies are well-motivated, they all assume \textit{perfect perception}, assuming precise knowledge of the trajectories of the surrounding pedestrians. In practice, such an assumption is unrealistic in high-density crowds. 

Additionally, these strategies tend to ignore an important structure in crowds - pedestrian groups~\cite{feng2014understanding}. Pedestrian motions are often coupled as a result of social grouping. As such, they can often be \textit{effectively} grouped by similarities in their motion characteristics. Poor modeling of the emergence of pedestrian groups may lead to aggressive and uncomfortable behavior for surrounding humans, violating social group space~\cite{wang2022gmpc}. Recognizing the importance of pedestrian groups, methods that factor pedestrian groups into social navigation have been proposed~\cite{katyal2022learning, medina2023human, luo2025gson, lu2025group}. These works share the common revelation that considering social groups yields safer navigation behavior and fewer group space intrusions for a mobile robot. However, these works also share the \textit{perfect perception} assumption, or rely on tracking modules to track pedestrians, which could be error-prone in crowded real-world settings. 

Interestingly, the Group-based Model Predictive Control (G-MPC) model proposed by Wang \etal ~\cite{wang2022gmpc} explored directly transferring the model to a simulated LiDAR scan setting. In other words, grouping can also be directly applied to LiDAR scans, thereby bypassing the need to track pedestrians first. However, the computation time required to process G-MPC is too long for practical deployment. 

\begin{figure}[t]
    \centering
    \includegraphics[width=0.99\linewidth]{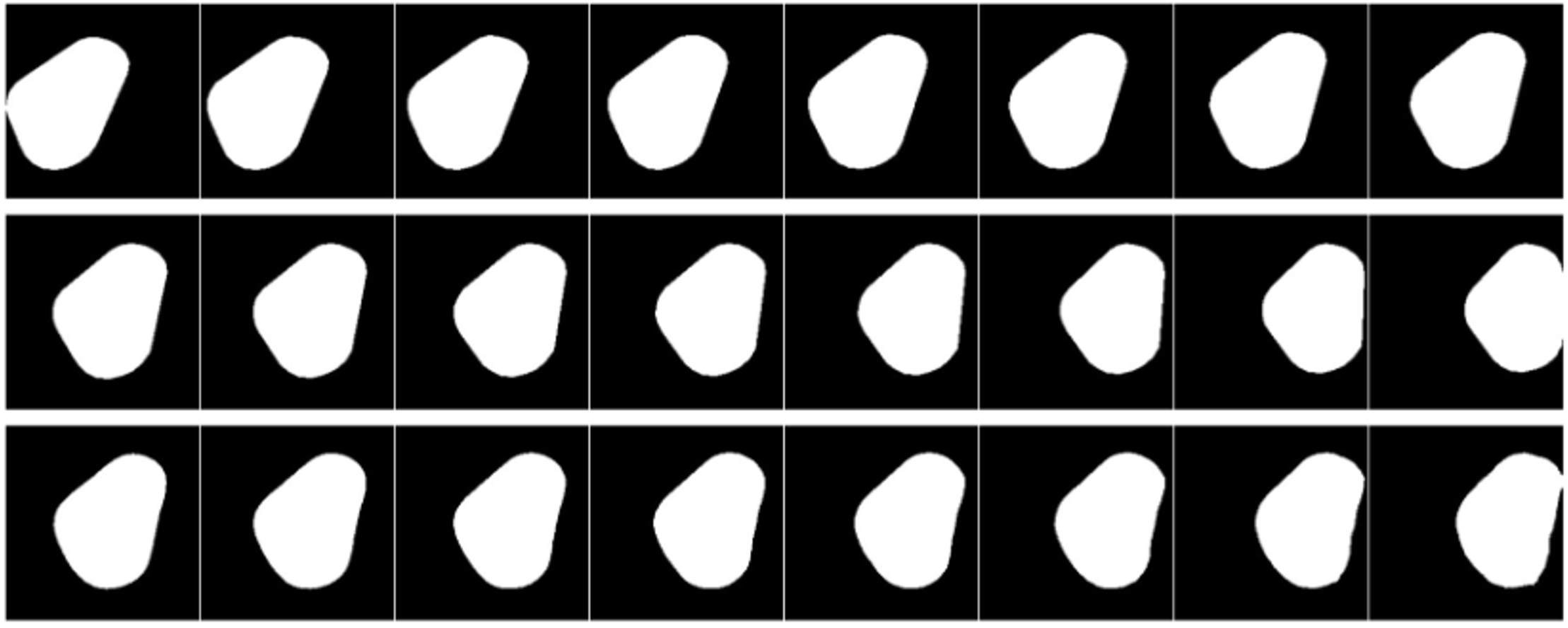}
    \caption{The top row is how an example group evolves in time. The middle row shows how the group evolves in the future according to the ground truth. The bottom row shows how the prediction model from~\cite{wang2022gmpc} predicts the future of this group. Despite the prediction model's small inaccuracies, the inaccuracies' impact on the overall navigation performance is small. This leads us to consider a much simplified group-based representation.}
    \label{fig:group_pred_exp}
    \vspace{-1em}
\end{figure}

Inspired by this work, we propose the visible edge-based representation for the practical deployment of a group-aware social navigation model, as shown in Fig.~\ref{fig:prop_new_gp}. We first conducted an in-depth analysis based on the framework of G-MPC and discovered that group prediction accuracies have a marginal impact on the overall social navigation performance (Fig.~\ref{fig:group_pred_exp}). This discovery led us to look for a solution that simplifies the group representations in G-MPC. When humans perceive crowds, we do not perform detailed tracking of individuals, but process groups. This observation is aligned with Gestalt theory from psychology~\cite{koffka2013principles}. We further introspected that when processing groups, we tend to focus on the visible parts of the groups. This inspired us to construct the visible edge-based representation.

We aim to demonstrate that our visible edge-based representation enables practical deployment of group-aware navigation without requiring pedestrian-tracking modules. Our main contributions are as follows. (1) We analyzed the effect of group prediction errors on social navigation performance and discovered that this effect is marginal. (2) We proposed a visible edge-based group representation that can be applied to LiDAR scan settings. We then integrated the representation into the framework of G-MPC~\cite{wang2022gmpc}. (3) We conducted experiments to show that our framework with our new representation performed similarly with the G-MPC model~\cite{wang2022gmpc}, but is significantly faster, which enables deployment on a real-world robot. (4) We deployed our framework on a real robot, demonstrating the model's practicality.
\section{Related Works}

\subsection{Social Navigation}
Social navigation has been an active area of research for decades and has gained popularity. Early model-based methods such as the social force model~\cite{helbing1995social} and ORCA~\cite{van2008reciprocal} leverage the notion of force and reciprocal velocity obstacles, respectively to model crowd interaction. More recently, HATEB~\cite{singamaneni2021human} leverages context-dependent costmaps. Sun \etal~\cite{muchen2025mixed} modeled social navigation as a probabilistic mixed strategy Nash equilibrium problem. These social navigation models mostly focus on individuals and have not explicitly considered groups. 

Another major branch of research involves, on the individual level, a combination of a planning module and a trajectory prediction model~\cite{heuer2023proactive, le2024social, mavrogiannis2017socially, trautman2015robot, kiss2021probabilistic}, where the planning module plans collision-free paths given the predictions from the prediction model that are sometimes coupled. Group-based interactions may be captured implicitly by the prediction models. However, such prediction models are usually trained on small real-world datasets such as ETH~\cite{pellegrini2009eth} and UCY~\cite{lerner2007ucy}, where there are only hundreds of instances of groups, and large pedestrian datasets that feature dense human crowds only appeared recently~\cite{wang2024tbd, martin2021jrdb} and have not been actively used by the prediction models for training. Therefore, the capability of the individual-based prediction modules to capture group interactions remains unexplored.

In line with the mainstream learning-based approaches in robotics, imitation learning~\cite{kretzschmar2016socially, raj2024rethinking} and DRL-based approaches~\cite{chen2017socially, liu2023intention, wang2023navistar, wu2023risk, chen2019crowd} gained significant traction. While group-based interactions may be implicitly captured by either the encoding of the observations or the policies directly, imitation learning-based approaches encounter similar dataset issues as the planning and prediction methods mentioned above, and DRL-based approaches are typically trained in Crowdsim~\cite{chen2019crowd} using ORCA~\cite{van2008reciprocal} to simulate pedestrians, which does not consider group formations.

Most of the literature discussed so far assumes \textit{perfect perception}, utilizing ground-truth pedestrian trajectory knowledge in simulation, or a human detection and tracking module in the real world. In this paper, we show that group-based representations can also be directly generated from sensors, such as LiDAR scans. Nevertheless, some recent works also take sensor readings as inputs to their models. Hirose \etal~\cite{hirose2023sacson, hirose2024selfi} used cameras for vision-based social navigation. Other works, such as~\cite{de2024spatiotemporal, liang2020realtime, xie2023drl}, integrated LiDAR or multi-model sensor inputs into a DRL framework. All these works have not explicitly considered groups, and, except~\cite{xie2023drl}, they are evaluated in limited crowd settings.

The aim of this paper is not to propose a state-of-the-art social navigation system, but to explore the practicality of group-based representations and the benefits they can bring to social navigation. We intend to explore the integration of visible edge-based representation into state-of-the-art social navigation frameworks in the future.

\subsection{Group-based representations}
Šochman and Hogg~\cite{vsochman2011knows} observed that 50-70\% of pedestrians walk in groups. Several works exist in the detection of groups, such as inferring probabilistic association of individual trajectories~\cite{pellegrini2010improving, feng2014understanding, chang2011probabilistic} or using graphical models with strong edges to indicate groups~\cite{chamveha2013social, khan2015detection, jahangard2023real}. Clustering is another common technique to model groups~\cite{solera2015socially, taylor2020robot, chatterjee2016performance}. Our group-based representation is compatible with any grouping method that assigns group memberships to the individuals. Grouping itself is not the focus of our contribution.

Group considerations have been empirically consolidated as being beneficial to many applications, such as trajectory prediction~\cite{xu2024dynamic, bae2022learning}. Social navigation similarly benefits from this. Early models, such as the extended social force model~\cite{moussaid2010walking} and O-space model~\cite{yang2019social} showed promising results from using groups. More recently, VLM-Social-Nav~\cite{song2024vlm} and GSON~\cite{luo2025gson} applied large vision-language models to reason about groups and plan socially acceptable paths. CoMet~\cite{sathyamoorthy2021comet} modeled group cohesion for better group reasoning. These works are promising, but they have not been evaluated in densely crowded scenarios. The work done by Katyal \etal~\cite{katyal2022learning}, Medina \etal~\cite{medina2023human}, Lu \etal~\cite{lu2025group} and Wang \etal~\cite{wang2022gmpc} is of particular relevance. However, these works, as well as the literature mentioned earlier in the paragraph, depend on pedestrian detectors, which may render them impractical in high-density crowds in the real world. In this paper, we built on the framework of~\cite{wang2022gmpc} and explored the practical deployment of group-based representations that operate directly on sensor inputs, such as LiDAR scans.
\section{Method}
\subsection{Problem Definition}

We define a workspace $\mathcal{W}\subseteq\mathbb{R}^2$ with $n$ dynamic pedestrians and a robot. The robot's state is $s\in\mathcal{W}$ and the pedestrian's state is $s^i\in\mathcal{W}, i\in\mathcal{N} = \{1,\dots, n\}$, where the states are the Cartesian coordinates. The robot navigates from an initial state $s_{0}$ to a goal state $s_{g}$ following a policy $\pi:\mathcal{W}^{n+1}\times\mathcal{U}\to\mathcal{U}$. The policy maps the observable state of the world $\boldsymbol{S} = s \cup_{i = 1:n} s^i$ to an action $u\in\mathcal{U}, \mathcal{U}\subseteq\mathbb{R}^2$, where the action space is velocity controls. They are Cartesian for holonomic robots and polar for differential drive robots. The robot does not know the pedestrians' goals $s^{i}_g$, or policies $\pi_i:\mathcal{W}^{n+1}\times\mathcal{U}^i\to\mathcal{U}^i$, $i\in\mathcal{N}$. We integrate our visible edge group representations into the policy $\pi$ to enhance the robot's navigation from $s_0$ to $s_g$. 

In the scenario of directly applying the policy $\pi$ to using LiDAR scans, we define the state of the scan points $s_t^i\in\mathcal{W}$ at time $t$ given $m_t$ scan point observations $i\in\mathcal{M}_t = \{1,\dots, m_t\}$. The scan points $\mathcal{M}_t$ do not have goals or policies, as they are decided by the pedestrians who are now invisible. $\mathcal{M}_t$ can change from time to time, and some of them belong to static obstacles.

\subsection{Augmented States}
\label{sec:aug_state}

We first define an augmented state for pedestrian or scan point $i$ as $q^i = (s^i, \theta^i, v^i)$, where $\theta^i\in [0, 2\pi)$ is the orientation and $v^i = ||u^i||\in \mathbb{R}^{+}$ is the speed of agent $i\in \mathcal{N}$ or scan point $i\in \mathcal{M}_t$. For pedestrians, under the \textit{perfect perception} assumption, we assume the orientation is aligned with their velocity $u^i$ extracted via finite differencing of their position over a timestep $\Delta t$. 

For LiDAR scan points, we also assume the orientation is aligned with their velocity. But since fast LiDAR flow estimation is an active area of research~\cite{kim2025flow4d}, we take inspiration from the TAGD features~\cite{de2024spatiotemporal}, perform tracking via low-level clustering using DBSCAN~\cite{ester1996dbscan}, and obtain the velocities for the scan points from the tracked clusters. Given the LiDAR scan positions $s^i, i\in \mathcal{M}_t$, we first cluster the scan points into $k\in\mathcal{K}_t$ clusters $C^k_t = \{i\in\mathcal{M}_t \mid c^i=k\}$. We then calculate the cluster center coordinates for each cluster $\boldsymbol{c}^k_t = \bar{\boldsymbol{s}}^k_t$, where $\boldsymbol{s}^k_t = \{s^i|i\in\mathcal{M}_t,c^i=k\}$. For each $s^i$, its velocity $u^i$ is obtained via finite differencing between its cluster center and the previous closest cluster center, $u^i=\min_l(||\boldsymbol{c}^k_t-\boldsymbol{c}^l_{t-1}||)/\Delta t, l\in\mathcal{K}_{t-1}, c^i=k$. Here we also filter out large static obstacles based on the size of the clusters. For future work, a more advanced LiDAR flow estimation method may be used to assign velocities for scan points. In the current state, the clustering-based method satisfies our requirement.

\begin{figure}[t]
    \centering
    \includegraphics[width=1.0\linewidth]{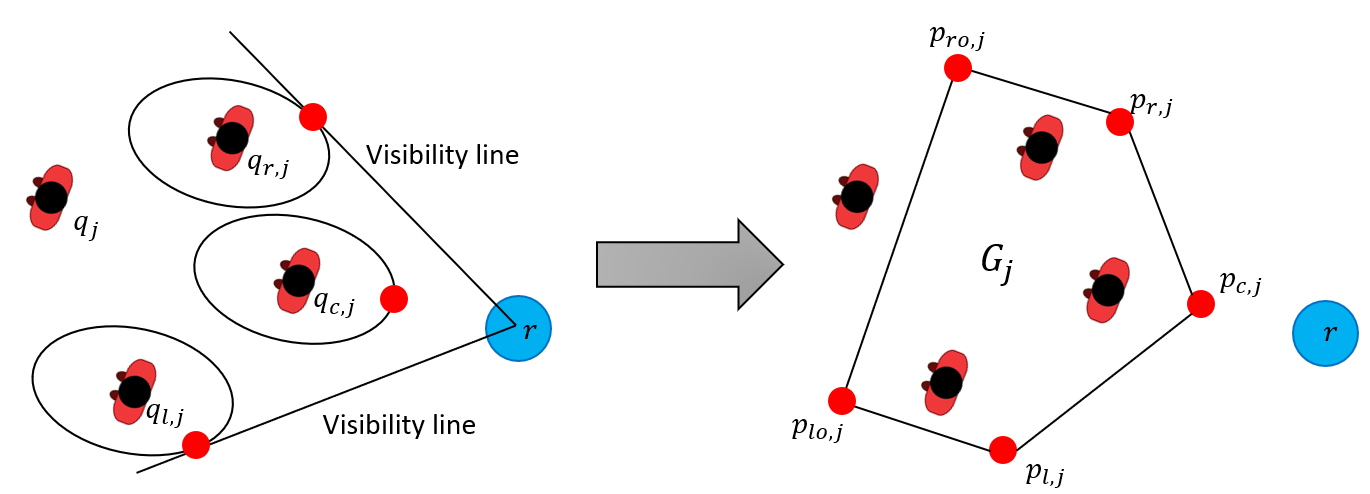}
    \caption{Demonstration of the construction of our visible edge-based representation. The blue circle is the robot. The robot first obtains the visible edges based on the proxemics of the pedestrians. Then it constructs a visible edge and an offset to form a pentagon. The pentagon is our proposed group-based representation and is defined by only three points, regardless of the size of the group.}
    \label{fig:prop_new_gp}
    \vspace{-1em}
\end{figure}

\subsection{Visible-Edge-Based Group Representation}
\label{sec:visible_edge}

We then define a group $j\in\mathcal{J}$ as a set $G^j = \{i \in \mathcal{N} \mid g^i = j \}$, or in the case of scan points $G^j = \{i \in \mathcal{M}_t \mid g^i = j \}$, and collect the set of all the groups in a scene into a set $\boldsymbol{G} = \{G^j\mid j\in\mathcal{J}\}$. We use DBSCAN again~\cite{chatterjee2016performance}, similar to G-MPC~\cite{wang2022gmpc}, to assign group memberships to agents. 

For each group $G^j$, $j\in\mathcal{J}$, we define a \emph{simplified social group space} as a geometric enclosure $\mathcal{G}^j$ around the agents or scan points of the group based on the visible edge of the group. The visible edge of the group is the edge of the group that is closest to the robot in the scene. 

As shown in Figure~\ref{fig:prop_new_gp}, we first define the visible edge of the group by identifying three key points: the point closest to the robot $p_{c, j}$; the point that is the leftmost visible point from the robot's perspective $p_{l, j}$; and the point that is the rightmost visible point from the robot's perspective $p_{r, j}$. To obtain $p_{c, j}$, $p_{l, j}$, and $p_{r, j}$, we first find the closest entity of the group $q_{c, j}$, the leftmost visible entity of the group $q_{l, j}$, and the rightmost visible entity of the group $q_{r, j}$. The group entity can either be a pedestrian or a LiDAR scan point. We then apply the egg-shaped personal space model $\mathcal{P}$ following G-MPC~\cite{wang2022gmpc} and establish a set of boundary points for each of the three entities $\mathcal{P}_{c, j}$, $\mathcal{P}_{l, j}$ and $\mathcal{P}_{r, j}$. Finally, $p_{c, j}$ is the closest point to the robot among $\mathcal{P}_c$. $p_{l, j}$ and $p_{r, j}$ are the leftmost and rightmost visible points from the robot's perspective among $\mathcal{P}_{l, j}$ and $\mathcal{P}_{r, j}$, respectively. After the three key points are identified, the visible edge consists of the two connecting line segments $\overline{p_{c, j}p_{l, j}}$ and $\overline{p_{c, j}p_{r, j}}$.

Next, we define a fixed offset behind the visible edge. Given $p_{l, j}$ and $p_{r, j}$, we apply an offset of width $d$ away from the robot and obtain the two offset points $p_{lo, j}$ and $p_{ro, j}$. We set this offset with length $d$ to account for occlusions close to the visible edge. We rely on the regeneration of group space to find new visible edges and offsets of the group for future timesteps if more occluded parts of the group emerge. With this, our updated group space is now a pentagon.
\begin{equation}\label{eqn:pentagon}
    \mathcal{G}^j=\mbox{Pentagon}(\{p_{l, j}, p_{c, j}, p_{r, j}, p_{ro, j}, p_{lo, j}\})\mbox{.}
\end{equation}
We collect the spaces of all groups in a scene into a set $\boldsymbol{\mathcal{G}} = \{\mathcal{G}^j\mid j\in \mathcal{J}\}$.

\subsection{Prediction Oracle}
\label{sec:oracle}

Our social group space is now defined by five points, of which only three points $\{p_{c, j}, p_{l, j}, p_{r, j}\}$ need to be tracked in order to predict the groups' future states. Compared to the convex hull-based group representations from G-MPC, our visible edge-based representation is more tractable. 

We track group states from time $t_h=t - h$ and make predictions up to time $t_f=t + f$. We collect a history of the trajectories for $\{p_{c, j}, p_{l, j}, p_{r, j}\}$ and obtain $\tau_{c, j} = p^{c, j}_{t_h:t}, \tau_{l, j} = p^{l, j}_{t_h:t}, \tau_{r, j} = p^{r, j}_{t_h:t}$. For LiDAR points, the tracking was curated using the low-level clustering-based method in Sec~\ref{sec:aug_state}. We perform the history gathering processing for every group and obtain $\boldsymbol{\mathcal{T}}_{t_h:t} = \{\tau_{c, j}, \tau_{l, j}, \tau_{r, j} \mid j\in \mathcal{J}\}$. With a set of history trajectories, we use Social-GAN~\cite{gupta2018sgan}, a popular trajectory prediction model, to predict the future trajectories of these key points of our visible edge-based representation.
\begin{equation}
    \boldsymbol{\mathcal{T}}_{t:t_f} = \mbox{SGAN}(\boldsymbol{\mathcal{T}}_{t_h:t})
\end{equation}
For each group $j$, we then obtain the predicted future key points $p^{c, j}_{t:t_f}, p^{l, j}_{t:t_f}, p^{r, j}_{t:t_f}$ from $\boldsymbol{\mathcal{T}}_{t:t_f}$. Next, we generate the offset points $p^{lo, j}_{t:t_f} $ and $ p^{ro, j}_{t:t_f}$ from the predicted key points, following the procedures in Sec.~\ref{sec:visible_edge}. Finally, we obtain the predicted future visible edge-based group space $\mathcal{G}^j_{t:t_f}$ from the completed key point sets.

We choose Social-GAN because it is a well-established trajectory prediction model that only requires positional trajectories as input. Models such as Sophie~\cite{sadeghian2019sophie} or Y-net~\cite{mangalam2021ynet} require image patches or semantic segmentation as additional input. Although better-performing trajectory prediction models exist~\cite{yuan2021agentformer, fu2025moflow}, Poddar \etal~\cite{poddar2023crowd} have shown that better trajectory prediction models only benefit downstream navigation performance marginally.

\subsection{Integration into MPC}

Integration of the group representation based on visible edges into a Model Predictive Control (MPC) framework is similar to G-MPC. At time $t$, we obtain an optimal control trajectory $\boldsymbol{u}^* = u^*_{1:K}$ of length $K$ by solving the following optimization problem\footnote{\label{ft:prior_work}Refer to G-MPC~\cite{wang2022gmpc} for more details}:
\vspace{-1em}
\begin{align}
\left(\boldsymbol{s}^*, \boldsymbol{u}^{*}\right) =&  \arg\min_{u_{1:K}} \sum_{k=1:K}^{K} \gamma^{k}J(s_{k+1}, \boldsymbol{\mathcal{G}}_{k+1}, s_T)\label{eq:cost}\\
 s.t.\: 
         & \boldsymbol{\mathcal{G}}_{2-h:1} \leftarrow \boldsymbol{\mathcal{G}}_{t_{h}:t}\label{eq:initgroups}\\
         & s_1 \leftarrow s_t\label{eq:initrobot}\\
         & \boldsymbol{\mathcal{G}}_{k+1:k_{f}} = \mathcal{O}(\boldsymbol{\mathcal{G}}_{k_{h}:k})\label{eq:updategroup}\\
         & u_k\in\mathcal{U}\\
         & s_{k+1} = s_k + u_k \cdot dt\label{eq:statetransition}
   \mbox{,}
\end{align}
where $\gamma$ is the discount factor, $\mathcal{O}(\boldsymbol{\mathcal{G}}_{k_{h}:k})$ is the group prediction oracle function, of which the procedures are described in Sec.~\ref{sec:oracle}, and $J$ is the cost function. The cost function uses a weighted sum of costs $J_g$ and $J_d$, representing penalties on distance to the robot's goal and proximity to groups:
\begin{equation}
    J(s_{k}, \boldsymbol{\mathcal{G}}_{k}, s_T) = \lambda J_{g}(s_{k}, s_T) +(1-\lambda)J_{d}(s_{k}, \boldsymbol{\mathcal{G}}_k)
    \label{eq:begin_detail_mpc}\mbox{,}
\end{equation}
where
\begin{equation}
    J_{d}(s_{k}, \boldsymbol{\mathcal{G}}_k) = \exp(-\mathcal{D}\left(s_{k+1}, \boldsymbol{\mathcal{G}}_k)\right)\mbox{,}
\end{equation}
\begin{equation}
    \mathcal{D}(s_{k}, \boldsymbol{\mathcal{G}}_k) = \min_{j\in \mathcal{J}} D\left(s_k - \mathcal{G}_k^j\right)
    \label{eq:end_detail_mpc}
\end{equation}

For visible edge-based representations, the differences are in how we define the distance function $D(s_k - \mathcal{G}_k^j)$. $D(s_k - \mathcal{G}_k^j)$ is the distance between the robot and the group $\mathcal{G}_k^j$.
\begin{equation}
    D(s_k - \mathcal{G}_k^j) = \min(\{||s_k - p||\}), p\in\overline{p_{l, j}p_{c, j}p_{r, j}p_{ro, j}p_{lo, j}}
\end{equation}
The visible edge-based group spaces are pentagons defined by five line segments, so we can directly apply geometric formulas to calculate the distance between the robot and a line segment five times per group.
\section{Evaluation}

\subsection{Implementation Details}
To allow fair comparison with G-MPC~\cite{wang2022gmpc}. The parameters related to grouping and MPC are the same as G-MPC. Similarly, for group predictions, we consider 8 steps of history and predict 8 steps into the future. We set the offset parameter $d$ of the visible edge group representations to be 1 meter. For the velocities of the LiDAR scan flows, in simulation experiments, we obtain the velocities based on the pedestrian's velocities, in real-world experiments, we set the DBSCAN clustering parameter to be $0.5m$ for the procedures in Sec.\ref{sec:aug_state}. For the SGAN prediction model~\cite{gupta2018sgan}, we applied the provided checkpoints and did not retrain the model. The checkpoint used depends on the subdataset used in the simulation evaluation scenario settings. In real-world experiments, the checkpoint for evaluating in ETH is used.

\subsection{Simulation Setup}

We follow a similar real-world dataset-based simulation evaluation setup as in G-MPC. We construct a ETH~\cite{pellegrini2009eth} (ETH, HOTEL) and UCY~\cite{lerner2007ucy} (ZARA1, ZARA2, UNIV) dataset-based simulation benchmark environment. For each scene, we define two navigation tasks: \textbf{Flow}, where the robot moves along the dominant crowd direction, and \textbf{Cross}, where the robot moves perpendicular to the crowd flow. We define a test region in the middle of the pedestrian flow, and generate challenging trials when at least 5 pedestrians are present in the test region. The robot's speed is $1.75 m/s$, $\Delta t=0.1s$, and we assume a collision radius of $0.5m$\footref{ft:prior_work}. We consider two evaluation settings: \textbf{Offline}, where pedestrians follow recorded trajectories according to the datasets without reacting to the robot, and \textbf{Online}, where pedestrians' movements are simulated by ORCA and can react to the robot dynamically. We refer to scenarios where the ground truth pedestrian information are available as \textbf{perfect perception} scenarios. Additionally, we consider a simulated 2D LiDAR scan setting, the \textbf{LiDAR} scenario. We model pedestrians as $0.5m$ radius circles and project rays out from the robot. If a ray hits a pedestrian circle, a simulated scan point is registered. The resolution of the simulated scan is $0.25$ degrees, and a uniform $\pm0.05m$ noise is injected into each simulated scan point's x and y coordinates.

\subsection{Baselines and Metrics}

We compare our proposed visible edge-based group MPC (\textbf{edge-MPC}) against multiple baselines spanning pedestrian-based MPC, convexhull group-based MPC with different prediction models, and an advanced deep reinforcement learning (DRL) based approach.

Pedestrian MPC baselines treat each pedestrian individually without group representations:
\textbf{ped-nopred} uses no prediction;
\textbf{ped-linear} uses constant-velocity linear prediction;
\textbf{ped-sgan} uses Social-GAN~\cite{gupta2018sgan} for trajectory prediction. Note that \textbf{ped-sgan} is only evaluated in the perfect perception setting, since Social-GAN requires individual pedestrian trajectories as inputs, which are not available in the simulated LiDAR setting. In the simulated LiDAR settings, the LiDAR scan points are treated as individual entities, so the baselines are renamed \textbf{lidar-nopred}, and \textbf{lidar-nopred}.

Group MPC baselines use convex hull-based group representations from G-MPC~\cite{wang2022gmpc} and variations with different prediction oracles:
\textbf{group-nopred} uses no prediction;
\textbf{group-linear} shifts groups by constant velocity prediction of the group shape centroids;
\textbf{group-sgan} uses Social-GAN to predict trajectories of the individual pedestrians first and then perform grouping on the predictions, also only evaluated in the perfect perception setting;
\textbf{group-conv} uses the convolutional image-sequence prediction model from G-MPC to predict group shapes. This is the G-MPC model.

DRL baseline: We include \textbf{crowdattn-rl}, a DRL-based method adapted from the intention-aware crowd navigation model with attention-based interaction graph by Liu \etal~\cite{liu2023intention}. This serves as a representative learning-based baseline and represents an advanced pedestrian-based approach. We evaluate it only in the perfect perception setting, as it requires individual pedestrian states as input.


We report four metrics, aggregated by averaging across all 10 scenarios (5 sub-datasets $\times$ 2 tasks):
Success Rate (SR~$\uparrow$), the portion of trials where the robot reaches the goal without collision or time-outs;
Minimum Distance (MinD~$\uparrow$, $m$), the minimum distance between the robot and any pedestrian during a trial;
Path Length (PL~$\downarrow$, $m$), the distance traveled by the robot; and
Computation Time (Time~$\downarrow$, $s$), the average per-step computation time. All evaluations are performed on a server machine with an Intel Xeon Silver 4214R CPU at 2.40GHz and an Nvidia GeForce RTX 3090 GPU.

\begin{table}[t]
\centering
\caption{Results in the \textbf{Perfect Perception} setting.}
\label{tab:perfect_perception}
\vspace{2pt}
\begin{tabular}{lcccc}
\toprule
Method & SR $\uparrow$ & MinD (m)  $\uparrow$ & PL (m) $\downarrow$ & Time (s) $\downarrow$ \\
\midrule
\multicolumn{5}{l}{\textit{Offline (pedestrians do not react)}} \\
\midrule
ped-nopred & 0.992 & 1.218 & 18.32 & {$<$}0.001 \\
ped-linear & 0.998 & 1.398 & 18.12 & {$<$}0.001 \\
ped-sgan & 0.963 & 1.329 & 17.79 & 0.051 \\
\midrule
group-nopred & 0.994 & 1.560 & 22.68 & 0.042 \\
group-linear & 0.998 & 1.677 & 22.46 & 0.332 \\
group-sgan & 0.993 & 1.659 & 22.87 & 0.389 \\
group-conv & 0.998 & 1.622 & 22.71 & 0.841 \\
edge-MPC (ours) & 0.971 & 1.578 & 22.29 & 0.190 \\
\midrule
crowdattn-rl & 0.982 & 0.809 & 14.98 & 0.074 \\
\midrule
\multicolumn{5}{l}{\textit{Online (pedestrians react via ORCA)}} \\
\midrule
ped-nopred & 1.000 & 1.300 & 18.51 & {$<$}0.001 \\
ped-linear & 0.999 & 1.464 & 18.14 & {$<$}0.001 \\
ped-sgan & 0.992 & 1.420 & 18.23 & 0.050 \\
\midrule
group-nopred & 0.993 & 1.672 & 23.56 & 0.037 \\
group-linear & 0.998 & 1.810 & 23.27 & 0.284 \\
group-sgan & 0.993 & 1.816 & 23.14 & 0.359 \\
group-conv & 0.998 & 1.795 & 23.21 & 0.571 \\
edge-MPC (ours) & 0.998 & 1.752 & 22.76 & 0.153 \\
\midrule
crowdattn-rl & 1.000 & 0.921 & 14.02 & 0.073 \\
\bottomrule
\end{tabular}
\vspace{-1em}
\end{table}

\subsection{Results}

\begin{table}[t]
\centering
\caption{Results in the \textbf{LiDAR} setting}
\label{tab:lidar}
\vspace{2pt}
\begin{tabular}{lcccc}
\toprule
Method & SR $\uparrow$ & MinD (m) $\uparrow$ & PL (m) $\downarrow$ & Time (s) $\downarrow$ \\
\midrule
\multicolumn{5}{l}{\textit{Offline (pedestrians do not react)}} \\
\midrule
lidar-nopred & 0.963 & 1.401 & 21.32 & {$<$}0.001 \\
lidar-linear & 0.971 & 1.567 & 21.09 & {$<$}0.001 \\
\midrule
group-nopred & 0.957 & 1.579 & 23.07 & 0.193 \\
group-linear & 0.976 & 1.705 & 23.12 & 1.397 \\
group-conv & 0.975 & 1.680 & 23.54 & 1.766 \\
edge-MPC (ours) & 0.928 & 1.629 & 23.85 & 0.131 \\
\midrule
\multicolumn{5}{l}{\textit{Online (pedestrians react via ORCA)}} \\
\midrule
lidar-nopred & 0.996 & 1.459 & 22.07 & {$<$}0.001 \\
lidar-linear & 0.996 & 1.668 & 21.39 & {$<$}0.001 \\
\midrule
group-nopred & 0.998 & 1.689 & 24.35 & 0.145 \\
group-linear & 0.998 & 1.840 & 23.69 & 1.097 \\
group-conv & 0.993 & 1.829 & 24.03 & 1.343 \\
edge-MPC (ours) & 0.996 & 1.762 & 23.61 & 0.115 \\
\bottomrule
\end{tabular}
\vspace{-1em}
\end{table}


\textbf{Ablation of group-based MPC with various prediction models.} From either the perfect perception scenarios or the LiDAR scenarios, and both the offline and online scenarios. The performance difference among group-linear, group-sgan, and group-conv are very negligible. Notably, they all perform at comparable or higher SR and larger MinD than group-nopred, meaning that the presence of the prediction models itself brings a sizable boost to the model's performance. As long as the prediction model can produce reasonable predictions, their prediction inaccuracies' impact on the navigation performance is marginal. This reaffirms our intuition to develop the simplified visible edge-based group representation model.

\textbf{The visible edge representation vs. other group methods.} Our proposed visible edge method achieves comparable SR, MinD, and PL values to the other group-based methods. To rigorously assess this claim, we perform Two One-Sided Tests (TOST) for equivalence comparing per-trial results of edge-MPC versus group-conv (G-MPC) across all scenarios. We define equivalence margins of $\pm0.2$m for MinD and $\pm2.0$m for path length. For MinD, equivalence is observed in all four conditions ($p_{\text{TOST}}<0.001$), demonstrating that the visible edge method maintains statistically equivalent safety margins to G-MPC. Path length equivalence is also observed in all conditions ($p_{\text{TOST}}<0.003$). For the success rate (margin $\pm0.03$), equivalence is observed in three of four conditions; the exception is the LiDAR offline setting, where edge has a lower SR (0.925 vs.\ 0.971, $p_{\text{TOST}}=0.923$), as pedestrians follow fixed trajectories and cannot avoid the robot. In the more realistic online settings, where pedestrians react, SR equivalence is demonstrated ($p_{\text{TOST}}<0.001$).

Among group-based methods with prediction, the visible edge method achieves the lowest computation time across all settings. In the perfect perception setting, edge-MPC takes 0.153--0.190s per step, compared to 0.332--0.389s for group-linear/group-sgan and 0.571--0.841s for group-conv (G-MPC). In the LiDAR setting, the advantage is even more pronounced: edge-MPC requires only 0.115--0.131s, while group-conv takes 1.343--1.766s and group-linear takes 1.097--1.397s. This shows that predicting only three key edge points is more efficient than predicting convex hull shapes. edge-MPC performs faster in LiDAR settings because much fewer groups are detected due to occlusion.


\textbf{Group-based methods vs. Pedestrian-based methods.} Group-based methods perform safer with larger MinD values, similar SR, and longer PL values. Compared to pedestrian-based methods, edge-MPC achieves $\sim$20--35\% higher MinD while maintaining comparable success rates, as group representations encourage the robot to act more conservatively and respect group spaces. Compared to the DRL baseline, edge-MPC nearly doubles the minimum pedestrian distance (1.578--1.752m vs.\ 0.809--0.921m), indicating substantially safer navigation. These safety gains come at the cost of longer paths ($\sim$22--24m vs.\ $\sim$18--22m for pedestrian-based methods and $\sim$14--15m for DRL), a direct consequence of taking wider detours.

\subsection{Qualitative Results}

\begin{figure}[t]
    \centering
    \includegraphics[width=1.0\linewidth]{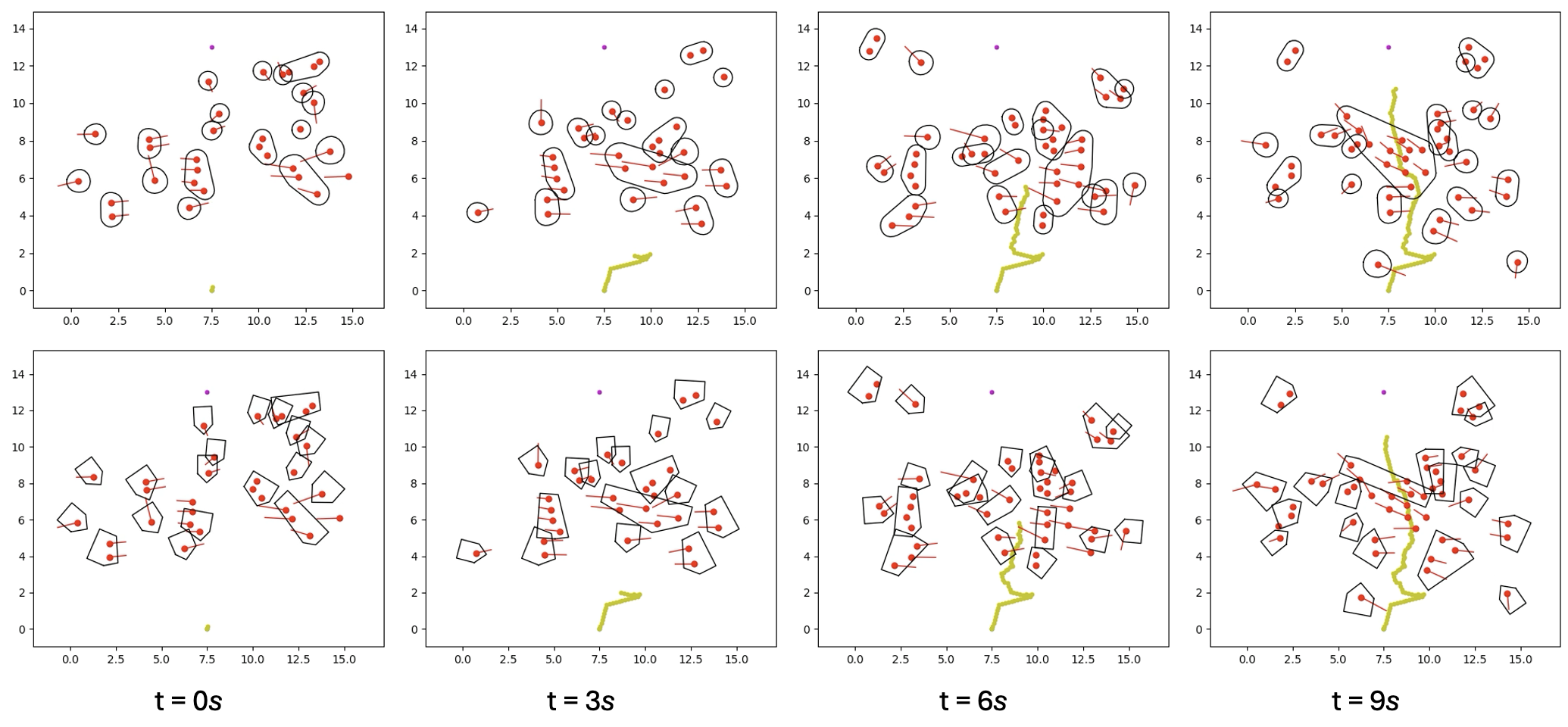}
    \caption{Qualitative examples of group-conv (top) vs. edge-MPC (bottom). Green trajectories show the robot trajectories. Red dots are pedestrians. Black enclosures are the group representations. Blue dots are the goals.}
    \label{fig:qual}
    \vspace{-1em}
\end{figure}

Fig.~\ref{fig:qual} demonstrates a qualitative comparison between G-MPC and our visible edge-based MPC. Despite the drastic changes to the group representations. The robot's overall navigation behavior and its decision making at key moments are very similar, as demonstrated by the similar trajectories that the robot has traversed in both cases. Similar to G-MPC, the visible edge-based robot's behavior is conservative. Please refer to the supplementary video for more examples.

\subsection{Real World Experiment}

\begin{figure*}[t]
    \centering
    \includegraphics[width=0.9\linewidth]{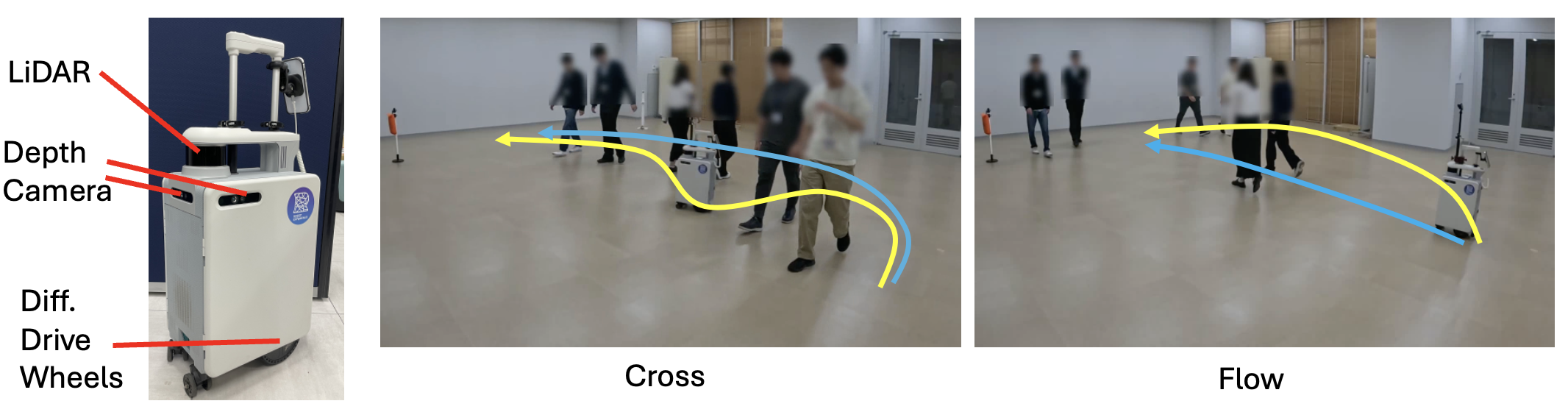}
    \caption{Left: The robot used in the real world experiment. Middle: The crossing scenario. Yellow trajectories illustrate edge-MPC's behavior. Blue trajectories illustrate crowdattn-RL's behavior. edge-MPC tried to walk behind the groups while crowdattn-RL often cut in front. Right: The flow scenario. edge-MPC performed a wide maneuver to avoid all the groups while crowdattn-RL followed the group in front and cut through the two incoming groups.}
    \label{fig:real_world}
    \vspace{-1em}
\end{figure*}

\begin{figure}[t]
    \centering
    \includegraphics[width=0.9\linewidth]{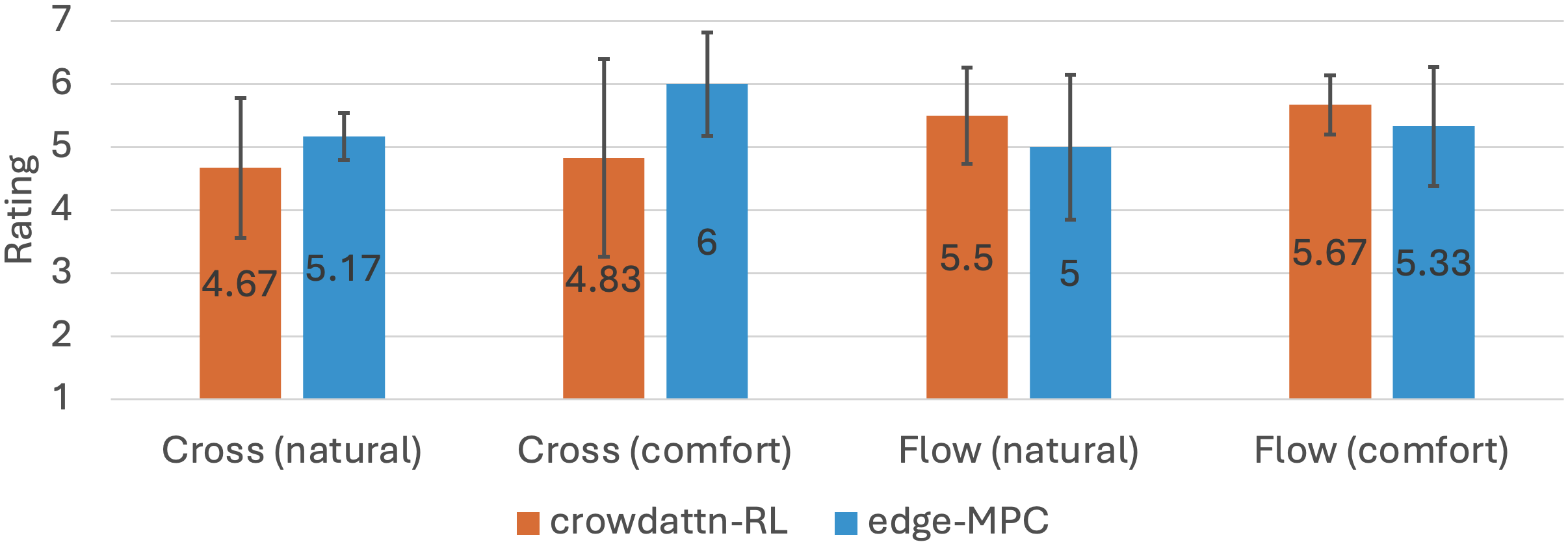}
    \caption{Ratings from the participants about the naturalness and comfort of the interaction they had with the robot during the trials.}
    \label{fig:rating}
    \vspace{-1em}
\end{figure}

We demonstrate the practicality of the visible edge group MPC framework by deploying it on a real-world robot. The robot is a custom robot as shown in Fig.~\ref{fig:real_world}. It is equipped with a Heisai XT32 3D LiDAR and three FRAMOS RealSense D455e depth cameras. The LiDAR is 3D, but we only read the middle scan ring to use the LiDAR as a 2D LiDAR. The cameras are pointed to the front, left, and right of the robot and cover about a 270-degree field of view. For models that require pedestrian tracking, We employ YOLOv4~\cite{bochkovskiy2020yolov4} coupled with SORT~\cite{bewley2016sort} to detect and track pedestrians from these cameras. We then read the depth values to project the coordinates of the tracked pedestrians onto the robot's map frame. All computations are processed over the robot's onboard Jetson AGX Orin.

We conducted the experiment in an open space spanning 10 meters x 9 meters. We recruited 6 college participants (4 men, 2 women) who were not familiar with robots in general. We asked the participants to form 3 groups of 2, and walk from one side of the room to the opposite side as naturally as possible. We created a cross scenario and a flow scenario similar to our simulation studies. As shown in Fig.~\ref{fig:real_world}, in the cross scenarios, two groups crossed paths with the robot from one side while one group crossed from the opposite side. In the flow scenarios, the robot followed one group while two groups walked in the opposite direction. For the robot, a few changes were made for its deployment in the real-world. For the MPC, we changed the sampling to the non-holonomic setting. The robot's max speed was set to $0.8m/s$, and max angular speed was $45^\circ/s$. We used the method in Sec.~\ref{sec:aug_state} to obtain LiDAR flow. We placed the robot so that it would cross paths with the participants. 

We tested two models in the real-world setting: \textbf{edge-MPC}, which is our visible edge-based group MPC approach, and \textbf{crowdattn-rl}, which requires the pedestrian detection and tracking module. In each of the cross and flow scenarios, the order of the models to experience was counterbalanced, and the participants were asked to try each model in each scenario three times. After each time, the three participant groups switch starting positions to experience a different perspective. After the three trials, they were asked to rate, on a scale of 1 to 7 (higher is better), the naturalness and comfort of the interaction experience with the robot. In total, for each rating, we collected 6 responses from 18 trials.

Our primary goal is to show that edge-MPC can be deployed practically on a real robot without pedestrian detections. The robot was able to consistently generate the group representations directly from LiDAR scans at around 20Hz and succeeded at reaching the goal without collisions. Qualitatively from Fig.~\ref{fig:real_world}, the edge-MPC behaves more conservatively and tries to detour behind the pedestrians in the cross scenario, whereas the crowdattn-RL approach often cuts in front of the pedestrians. In the flow scenario, the crowdattn-RL approach tends to follow the group in front and pass the two approaching groups by cutting through in the middle, while the edge-MPC robot favors performing a wide maneuver to avoid the groups altogether. According to the ratings from Fig.~\ref{fig:rating}, participants preferred visible edge-based group MPC both in terms of naturalness and comfort over the crowdattn-RL method in the crossing scenario. However, the edge-MPC was rated lower both in terms of naturalness and comfort in the flow scenario, despite the fact that the robot behaved more conservatively, avoided the groups altogether, and kept a larger distance from the groups. Conversations with the participants revealed concern about the difficulty in predicting the robot's motion during the wide maneuver.
\section{Conclusion}

We discovered that group prediction accuracies have marginal impact on downstream navigation performance, so we proposed a method of defining visible edges and generating simplified group space representations based on these visible edges. This formulation can be applied to LiDAR scans and does not require tracking of pedstrians. We additionally adopted a trajectory prediction model to predict groups that contains many fewer parameters than G-MPC's image sequence predictor model as our oracle. We show that with these improvements, we are able to achieve much faster computation speeds while maintaining similar levels of navigation performance, allowing us to deploy this model on a real robot.

Many improvements can still be applied to our framework. Motion predictability should be considered for future work~\cite{taylor2022observer}. Possible MPC improvements are applicable, and it is possible to switch to a reinforcement learning-based navigation framework that supports LiDAR scans as inputs. Although unlikely as suggested by~\cite{poddar2023crowd}, better trajectory prediction models such as Agentformer~\cite{yuan2021agentformer} or MoFlow~\cite{fu2025moflow} can be explored and tested to see if they offer significant improvements in navigation performance.





\section*{ACKNOWLEDGMENTS}
This work was supported by the AI Suitcase consortium and the staff members of the National Museum of Emerging Science and Innovation. Part of the visible edge-based representation is also based on the author's thesis.~\cite{wang2023thesis}

{
\bibliographystyle{bib/IEEEtran}
\bibliography{bib/IEEEabrv}
}

\end{document}